\documentclass[conference]{IEEEtran}
\IEEEoverridecommandlockouts
\usepackage{cite}
\usepackage{amsmath,amssymb,amsfonts}
\usepackage[ruled,vlined]{algorithm2e}
\usepackage{graphicx}
\usepackage{textcomp}
\usepackage{booktabs} 
\usepackage{tabularx} 
\usepackage[dvipsnames]{xcolor}
\usepackage{xcolor}
\usepackage{caption}
\usepackage{subcaption}
\usepackage{tabularray}   
\usepackage{booktabs}

\begin{document}


\title{Image Classification using Fuzzy Pooling in Convolutional Kolmogorov-Arnold Networks}

\author{\IEEEauthorblockN{Ayan Igali ,
Pakizar Shamoi\IEEEauthorrefmark{1}}
\IEEEauthorblockA{School of Information Technology and Engineering \\
Kazakh-British Technical University\\
Almaty, Kazakhstan\\
Email: 
\IEEEauthorrefmark{1}p.shamoi@kbtu.kz,
}
}

\maketitle

\begin{abstract}


Nowadays, deep learning models are increasingly required to be both interpretable and highly accurate. We present an approach that integrates Kolmogorov-Arnold Network (KAN) classification heads and Fuzzy Pooling into convolutional neural networks (CNNs). By utilizing the interpretability of KAN and the uncertainty handling capabilities of fuzzy logic, the integration shows potential for improved performance in image classification tasks. Our comparative analysis demonstrates that the modified CNN architecture with KAN and Fuzzy Pooling achieves comparable or higher accuracy than traditional models. The findings highlight the effectiveness of combining fuzzy logic and KAN to develop more interpretable and efficient deep learning models. Future work will aim to expand this approach across larger datasets. 


\end{abstract}


\section{Introduction}


In the field of computer vision, particularly in image classification tasks, the traditional approach has predominantly relied on multilayer perceptions (MLP). Over time, these gave way to more advanced architectures such as Convolutional Neural Networks (CNNs) \cite{alzubaidi2021review} and Vision Transformers (ViTs) \cite{visiontransformers}. However, one aspect that has remained consistent is the use of fully connected neural networks in the final layers of these classification models.

The main issues of traditional feedforward neural networks are complex interpretability and the number of parameters, which require a huge amount of computational power and memory space \cite{zhang2021survey}. A novel approach in machine learning, Kolmogorov-Arnold Network (KAN), which can become an alternative to traditional MLP architecture \cite{kan_paper}. KAN employs learnable activation functions inside represented with splines and "neurons", or MLPs, outside, which results not only in the learning of features but also in the optimization of learned features for better performance. The interpretability of KAN is demonstrated by the transparency of each activation depending on its magnitude, so it is immediately clear which input variables are important without the need for feature assignment.


Another approach toward interpretable and explainable AI models is Fuzzy Logic and Sets \cite{fernandez2019evolutionary, alonso2021toward}. Currently, the majority of data processed in information systems is of a precise nature \cite{precise}. The advantage of fuzzy logic is that it is close to the way a person thinks about a certain thing, which cannot be limited to 0 or 1 but is a range from 0 to 1. This helps deal with the uncertainties in the values of the features on which the model can be trained.  One of the novel approaches in image classification that incorporate fuzzy logic is the Type-1 Fuzzy Pooling layer, which aims to better preserve the original feature maps and copes with uncertainties of feature values \cite{fuzzy_pooling}. 

By integrating KAN with fuzzy pooling into a CNN, we can potentially leverage the strengths of both approaches to address issues of interpretability, memory usage efficiency, and uncertainty management in image classification problems. This integration represents a significant advance in the development of AI models that are more closely related to human perception and practical computational constraints.

The main contributions of this paper are:
\begin{itemize}
\item  We introduce a novel neural network architecture by integrating a KAN classification head and fuzzy pooling into standard CNNs. 
\item Comparative evaluation of fuzzy pooling techniques compared to traditional pooling methods like max and average pooling. Specifically, we analyze how different pooling strategies affect the performance of CNNs enhanced with a KAN head on standard image classification datasets CIFAR-10, MNIST, and FashionMNIST.
\end{itemize}
The paper is structured as follows. Section I is this Introduction. Section II presents an overview of the literature in related areas. Our Methodology is described in Section III. Next, we present the experimental results in Section IV. Finally, Section V provides concluding remarks and ideas for future improvements.

\section{Related Works}
\subsection{MLP architecture and their limitations.}

MLP, also called feedforward neural network is a fundamental architecture in the field of artificial intelligence and has become an essential backbone for a lot of machine learning applications \cite{goodfellow2016, lecun2015}. The primary advantage of MLPs is their universal approximation capability which can approximate any continuous function with enough amount of neurons in the hidden layers \cite{cybenko1989}. 

The applications of feedforward neural networks are limitless. They excel in tasks such as image classification and object detection, regression and prediction, machine translation, audio processing, and the generation of new content, whether it be a text, image, or video \cite{goodfellow2016}. 

Despite their power, these models, particularly MLPs, are often viewed as black boxes due to their complex nature \cite{lecun2015}. Furthermore, these deep learning models require extensive data to learn patterns effectively and generalize well \cite{schmidhuber2015}. Another significant limitation is the enormous computational power needed to train these models, a considerable drawback given that the latest large language models have parameters numbering in the hundred billion, and this requirement for high computational resources poses challenges, especially as model sizes continue to grow exponentially\cite{attention2017, kaplan2020}. 


\subsection{Kolmogorov-Arnold Networks}


The KAN is a promising alternative to MLPs for image classification, offering improved accuracy and interpretability \cite{kan_paper}. It is based on the Kolmogorov-Arnold representation theorem, which decomposes a multivariate function into an interior and an outer function \cite{J2020Kolmogorov}. The KAN has no linear weights, with every weight parameter replaced by a univariate function \cite{kan_paper}. It can be trained with fast and simple procedures, making it a practical choice for image classification \cite{Yevgeniy2005Neuro}. The KAN's performance has been demonstrated in various applications, including the classification of complex biological aging images \cite{T2014Classification}.

A range of other studies have explored the application of the Kolmogorov theorem in neural network training. The  Chebyshev Kolmogorov-Arnold Network introduced a novel approach for approximating complex nonlinear functions \cite{SS2024Chebyshev}. \cite{Joan2012Invariant} and \cite{T2014Classification} both focus on image classification, with \cite{Joan2012Invariant} discussing wavelet scattering networks and \cite{T2014Classification} proposing the use of fuzzy Kolmogorov-Sinai entropy for classifying biological aging images.  \cite{Yevgeniy2005Neuro} introduced a Neuro-Fuzzy Kolmogorov's Network for time series prediction and pattern classification. \cite{J2020Kolmogorov} revisits the Kolmogorov-Arnold representation theorem, discussing its potential application in deep neural networks.

\subsection{Fuzzy Logic in Neural Networks/Deep Learning}


Recent research has explored incorporating fuzzy logic into CNNs to enhance image classification performance. Several studies have proposed novel fuzzy pooling methods to replace traditional max or average pooling layers \cite{fuzzy_pooling}, \cite{Teena2019Fuzzy}, \cite{S2023Image}. These approaches aim to better handle uncertainty in feature maps and improve dimensionality reduction. Other researchers have investigated using the Choquet integral as a pooling function \cite{C2018Using} and introducing fuzzy layers into deep learning architectures \cite{Stanton2019Introducing}. Some studies have developed convolutional neural networks combining CNNs with fuzzy logic, incorporating fuzzy self-organization layers for data clustering \cite{V2019Multiclass}, \cite{K2018Convolutional}. Additionally, a deep neuro-fuzzy network with fuzzy inference and pooling operations was proposed \cite{Omolbanin2019Deep}, which is based on the Takagi-Sugeno-Kang fuzzy model. These fuzzy-based approaches have shown promising results in improving classification accuracy and handling uncertainty in image data.

\section{Methods}
\subsection{Data}

The current paper employs CIFAR-10, MNIST, and FashionMNIST datasets, essential in image classification research because they provide standardized benchmarks \cite{cifar10, lecun1998mnist, fashionmnist}. This consistency allows researchers to compare and validate different models and techniques, regardless of their complexity. Table \ref{tab:datasets} presents the basic information about these datasets.

\begin{table}[htb]
\centering
\caption{Summary of datasets used in research}
\label{tab:datasets}
\begin{tabular}{@{}p{2cm}p{1.5cm}p{2cm}p{2cm}@{}}
\toprule
\textbf{Dataset} & \textbf{\# of Images} & \textbf{Description}                         & \textbf{\# of Classes} \\ \midrule
CIFAR-10              & 60,000                    & 32x32 color images in 10 classes             & 10                         \\
MNIST                 & 70,000                    & Handwritten digits 0-9                      & 10                         \\
Fashion-MNIST         & 70,000                    & Fashion articles images     & 10                         \\ \bottomrule
\end{tabular}
\end{table}



\begin{figure}[htbp]
    \centering
        \includegraphics[width=\linewidth]{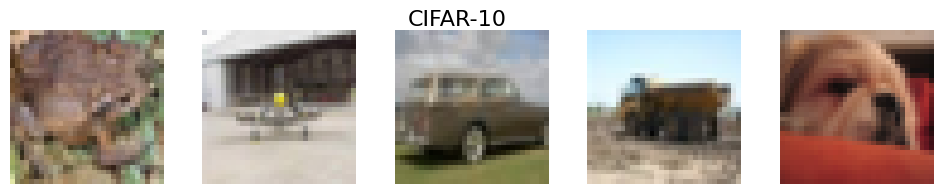}
        
        \includegraphics[width=\linewidth]{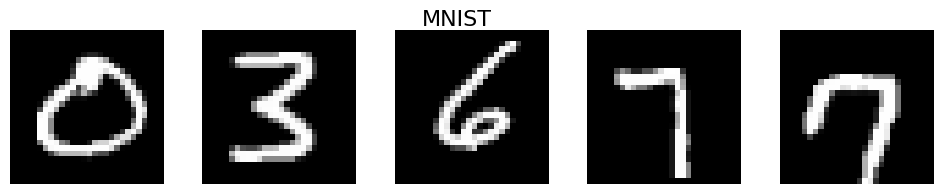}
        \includegraphics[width=\linewidth]{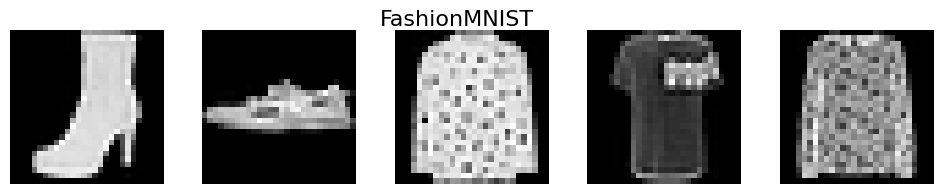}
    \caption{Example of images from datasets.}
    \label{image_example}
\end{figure}

\subsubsection{CIFAR-10}
CIFAR-10 is a collection of 60,000 32x32 color images of different classes, specifically airplanes, cars, birds, cats, deer, dogs, frogs, horses, ships, and trucks (each class has 6000 images).  
\subsubsection{MNIST}
MNIST is a dataset of 70,000 handwritten digit images 28x28 in grayscale. It is usually divided into 60,000 for training and 10,000 for testing.
\subsubsection{FashionMNIST} FashionMNIST is another commonly used dataset in image classification tasks, which was introduced as an alternative to the original MNIST dataset. FashionMNIST contains 70,000 28x28 images in a grayscale of 10 different fashion products: T-shirts, trousers, pullovers, dresses, coats, sandals, shirts, sneakers, bags, and ankle boots.

Figure \ref{image_example} shows image samples from employed datasets. No image augmentation or transformation was used to clarify model performance besides resizing MNIST and FashionMNIST datasets to fit the input requirements.

\subsection{KAN Layer}
The proposed approach integrates KAN into the existing CNN architectures to replace MLP as classification head in the last layers \cite{kan_paper}. The major difference between KAN and MLPs is flexible and trainable activation functions on the network's edges, represented with a spline function. 

The Kolmogorov-Arnold Networks can be described by the following formula \cite{kan_paper}:
\begin{equation}
f(x) = \sum^{2n+1}_{q=1}\Phi_{q}(\sum^{n}_{p=1}\phi_{q,p}(x_{p}))  
\end{equation}
Where $\Phi_q$ is transformations and $\phi_{q,p}(x_{p})$ is activation function which is sum of basis function $b(x)$ and spline function:

\begin{equation}
    \phi(x) = w_{b}b(x) + w_{s}spline(x)
\end{equation}

$b(s)$ is set as:

\begin{equation}
    b(x) = silu(x) = \frac{x}{1 + \exp^{-1}}
\end{equation}
where silu is Sigmoid Linear Unit activation function \cite{silu_article}.

$spline(x)$ is parameterized as a linear combination of B-splines such that 

\begin{equation}
    spline(x) = \sum_{i}c_{i}B_{i}(x)
\end{equation}

where $ c_i$ is trainable. $w_b$ and $w_s$ are trainable factors that can better control the activation function's overall magnitude.

The formula for deep KAN is \cite{kan_paper}:
\begin{equation}
    \text{KAN(x)} = (\Phi_3 \circ \Phi_2 \circ \Phi_1)(x)
\end{equation}

\subsection{Fuzzy Pooling Layer}

Due to the uncertainty of feature values of image data with possible noise, our proposed approach uses the Type-1 Fuzzy Pooling subsampling layer. This approach is distinct from traditional pooling methods like max or average pooling by incorporating fuzzy logic principles to handle data uncertainty effectively. The key operations of Type-1 Fuzzy Pooling are:


\begin{itemize}
    \item \textit{Fuzzification. }Converting crisp inputs into fuzzy sets allows the pooling layer to handle data uncertainty.
    \item \textit{Aggregation.} After fuzzification, the fuzzy pooling layer aggregates the fuzzified values within each pooling window using fuzzy logic operations. This involves methods like the fuzzy algebraic sum, which considers the contributions of all elements within the window rather than selecting a single maximum or averaging.
    \item \textit{Defuzzification.} The final step is where the aggregated fuzzy data is converted back into crisp outputs.

\end{itemize}

Let us consider a set of three fuzzy sets defined by 
\begin{equation}
    \tilde{y}_v = \left\{\langle x, \mu_v(x) \rangle \mid x \in E\right\},
    v = 1, \ldots, V
\end{equation}
for $V=3$ \cite{fuzzy_pooling}.

Let $p^{n}_{i,j}$ to be an element of volume patch $p$ at depth $n$ and position $i,j$ where $i = 1, \ldots, k, j = 1, \ldots, k$ and $k$ is pooling window size. The triangular membership function $\mu_1, \mu_2, \mu_3$ are defined for fuzzification of patches, which can be expressed as \cite{fuzzy_pooling}:
\begin{equation}
\mu_1(p^n_{i,j}) = 
\begin{cases}
0 & \text{if } p^n_{i,j} > d \\
\frac{d-p^n_{i,j}}{d-c} & \text{if } c \leq p^n_{i,j} \leq d \\
1 & \text{if } p^n_{i,j} < c
\end{cases}
\end{equation}
where $d= \frac{r_{max}}{2}$ and $c=\frac{d}{3}$,

\begin{equation}
\mu_2(p^n_{i,j}) = 
\begin{cases}
0 & \text{if } p^n_{i,j} \leq a \text{ or } p^n_{i,j} \geq b \\
\frac{p^n_{i,j}-a}{m-a} & \text{if } a < p^n_{i,j} \leq m \\
\frac{b-p^n_{i,j}}{b-m} & \text{if } m < p^n_{i,j} < b
\end{cases}
\end{equation}
 where $a=\frac{r_{max}}{4} ,m = \frac{r_{max}}{2}$ and $b = m + a$,

\begin{equation}
\mu_3(p^n_{i,j}) = 
\begin{cases}
0 & \text{if } p^n_{i,j} < r \\
\frac{p^n_{i,j}-r}{q-r} & \text{if } r \leq p^n_{i,j} \leq q \\
1 & \text{if } p^n_{i,j} > q
\end{cases}
\end{equation}

where $r = \frac{r_{max}}{2}$ and $q=r + \frac{r_{max}}{4}$. 

As studies suggest that it has been shown, the value of $r_{max}$ = 6 helps the network learn sparse features earlier \cite{fuzzy_pooling, mobilenet}. 

For each patch $p^n$, $n = 1, ..., z, $ a fuzzy patch $\pi^n_v$ is calucated as \cite{fuzzy_pooling}:
\begin{equation}
\pi^n_v = \mu_v(p^n)
\end{equation}

Aggregate the fuzzified values in each pooling window using the fuzzy algebraic sum \cite{Zimmermann2001}:



\begin{equation}
s_{\pi_v}^n = \overset{k}{\dot{\sum_{i=1}}} \overset{k}{\dot{\sum_{j=1}}} \pi_{v_{i,j}}^n, \quad n = 1, \ldots, z
\end{equation}


where $s^n_{\pi_{v}}$ score score quantifying the overall membership of $p^n$ to $\tilde{y}_v$. Based on these scores for each patch $p$ a new fuzzy volume patch $\pi'$ is created by selecting the spatial fuzzy patches $\pi^n_v$, $v=1,\ldots, V$ that have largest score $s^n_{\pi_{v}}$

\begin{equation}
\pi' = \left\{\pi'{_v}{^n} = \pi_v^n \mid v = argmax(s_{\pi_v}^n), n = 1,2,\ldots,z\right\}
\end{equation}

After the patch with the highest certainty is selected, the dimension of each patch is reduced by defuzzification by using the center of gravity (CoG)

\begin{equation}
p'^n = \frac{\sum_{i=1}^k \sum_{j=1}^k (\pi_{i,j}^n \cdot p_{i,j}^n)}{\sum_{i=1}^k \sum_{j=1}^k \pi_{i,j}^n}, n = 1 \ldots z
\end{equation}

Where $p' = \{p'^n \mid n = 1,2,\ldots,z\}$.

\subsection{Proposed Approach}

As a proposed approach, we suggest a hybrid system of CNNs with a Fuzzy Pooling layer for sampling feature maps and KAN as the classification head on the network's last layer instead of traditional MLP. This methodology is not tied to exact model architecture but a new approach for image classification tasks. The ability of Fuzzy Pooling to extract valuable features from uncertain and noisy feature maps \cite{fuzzy_pooling} incorporated with the adaptivity of KAN \cite{kan_paper} can result in more efficient models in terms of memory usage because KANs need fewer parameters to match the performance of MLP. 
\begin{figure*}[ht!]
    \centering
        \includegraphics[width=\textwidth]{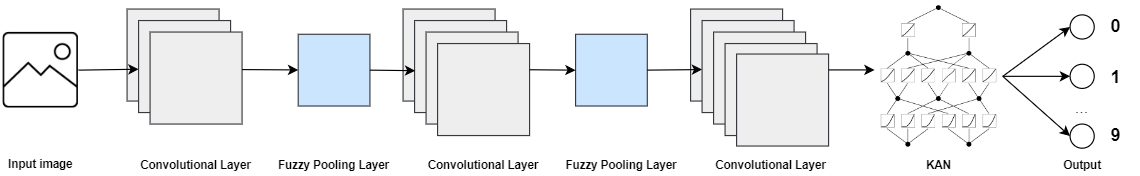}
    \caption{Schematic representation of the proposed architecture}
    \label{methodology_fig}
\end{figure*}

\begin{algorithm}
\SetAlgoLined
\KwResult{Trained CNN model with KAN and Fuzzy Pooling}
\textbf{Input:} Image dataset $D$\;


\SetKwFunction{FPreprocessData}{PreprocessData}
\SetKwFunction{FInitializeCNN}{InitializeCNN}
\SetKwFunction{FForwardPass}{ForwardPass}
\SetKwFunction{FBackwardPass}{BackwardPass}
\SetKwFunction{FTrainModel}{TrainModel}

\SetKwProg{Fn}{Function}{:}{}

\SetKwProg{Fn}{Function}{:}{}
\Fn{\FInitializeCNN{}}{
    Initialize layers with standard CNN architecture\;

    Replace pooling layers with Fuzzy Pooling\;
    
    Replace the final classification layer with KAN\;
    
}

\SetKwProg{Fn}{Function}{:}{}
\Fn{\FForwardPass{$x$}}{
    \For{each layer $l$ in CNN}{
    $x \gets \text{FuzzyPooling}(x)$\;
    
    $x \gets \text{Activate}(\text{Convolve}(x, weights_l))$\;
    }
    $x \gets \text{KANClassify}(x)$\;
    \KwRet $x$\;
}

\SetKwProg{Fn}{Function}{:}{}
\Fn{\FBackwardPass{$\text{loss}, \text{output}, \text{target}$}}{
    Compute gradients w.r.t loss\;
    
    Update all weights using AdamW optimizer \;
}

\SetKwProg{Fn}{Function}{:}{}
\Fn{\FTrainModel{$epochs, D$}}{
    \For{$epoch = 1$ to $epochs$}{
        \For{each batch $b$ in $D$}{
            $output \gets \text{ForwardPass}(b)$\;
            
            $loss \gets \text{ComputeLoss}(output, \text{true labels of } b)$\;
            
            \text{BackwardPass}(loss, output, true labels of $b$)\;
        }
    }
    Evaluate model on validation set\;
}

\caption{CNN with KAN Classification Head and Fuzzy Pooling}

\end{algorithm}

For our research, we decided to continue the practice of the authors of Type-1 Fuzzy Pooling and use LeNet architecture to evaluate the performance of the proposed approach \cite{fuzzy_pooling}.  Figure \ref{methodology_fig} presents our approach and illustrates how Fuzzy Pooling and KAN layers are implemented in LeNet architecture \cite{lenet_1989}.




\section{Results}
\subsection{Experiment Setup}
We trained a proposed modified LeNet model featuring Fuzzy Pooling layers and a KAN classification head on the CIFAR-10 (50,000 training and 10,000 test images), MNIST (60,000 training and 10,000 test images), and FashionMNIST (60,000 training and 10,000 test images) datasets. The model underwent training over 10 epochs using the AdamW optimizer with a learning rate 0.001 and a batch size of 32.

For comparative analysis, we trained several variants of the LeNet model:
\begin{itemize}
    \item LeNet with a KAN layer and traditional Max and Average Pooling subsampling layers.
    \item LeNet is equipped with an MLP in the final layer, Fuzzy, Max, and Average Pooling subsampling layers configured identically to the proposed model.
\end{itemize}

These configurations allowed for a comprehensive evaluation of the modified model's performance relative to standard pooling methods and network architectures.
\subsection{Performance Evaluation}

We employ measures such as Accuracy, Precision, Recall, and the \(F_1\) score to assess our classification model. 


\textbf{Accuracy}. The proportion of true results among the total number of cases examined.
\begin{equation}
    \text{Accuracy} = \frac{TP + TN}{TP + TN + FP + FN}
\end{equation}
where \(TP\) (true positives) and \(TN\) (true negatives) represent correct predictions, while \(FP\) (false positives) and \(FN\) (false negatives) represent incorrect predictions.

\textbf{Precision}. The proportion of positive identifications that were actually correct.
\begin{equation}
    \text{Precision} = \frac{TP}{TP + FP}
\end{equation}

\textbf{Recall} (also known as sensitivity). The proportion of actual positives that were correctly identified.
\begin{equation}
    \text{Recall} = \frac{TP}{TP + FN}
\end{equation}

\textbf{F1 Score}. The harmonic mean of precision and recall.  A higher \(F_1\) score indicates a closer approximation of Precision and Recall.

\begin{equation}
    \text{F1 Score} = 2 \cdot \frac{\text{Precision} \cdot \text{Recall}}{\text{Precision} + \text{Recall}}
\end{equation}

\begin{table*}
\centering
\resizebox{1\textwidth}{!}{%
\begin{tabular}{l|cccc|cccc|cccc}
\toprule
\multicolumn{1}{l}{} & \multicolumn{4}{c}{\textbf{CIFAR-10}} & \multicolumn{4}{c}{\textbf{FashionMNIST}} & \multicolumn{4}{c}{\textbf{MNIST}} \\
\hline
\textbf{Methods} & {Accuracy} & {Precision} & {Recall} & {F1-Score} & {Accuracy} & {Precision} & {Recall} & {F1-Score} & {Accuracy} & {Precision} & {Recall} & {F1-Score} \\
\hline
MLP + Average Pooling & 65.66 & 0.66 & 0.66 & 0.66 & 89.48 & 0.89 & 0.90 & 0.89 & 98.89 & 0.99 & 0.99 & 0.99 \\
MLP + Max Pooling & 66.69 & 0.67 & 0.67 & 0.66 & 89.81 & 0.90 & 0.90 & 0.90 & 98.88 & 0.99 & 0.99 & 0.99 \\
MLP + Type-1 Fuzzy Pooling & 65.60 & 0.66 & 0.66 & 0.65 & 89.22 & 0.89 & 0.90 & 0.89 & 98.51 & 0.98 & 0.99 & 0.98 \\
KAN + Average Pooling & 63.39 & 0.63 & 0.64 & 0.63 & 88.84 & 0.89 & 0.89 & 0.89 & 98.60 & 0.99 & 0.99 & 0.99 \\
KAN + Max Pooling & 66.92 & 0.67 & 0.67 & 0.67 & 89.12 & 0.89 & 0.89 & 0.89 & 98.88 & 0.99 & 0.99 & 0.99 \\
\textbf{KAN + Type-1 Fuzzy Pooling} &\textbf{67.06} & \textbf{0.67} &\textbf{0.67} & \textbf{0.67} &\textbf{89.88} &\textbf{0.90} & \textbf{0.90 }& \textbf{0.90 }& \textbf{98.91} &\textbf{0.99} & \textbf{0.99 }& \textbf{0.99} \\
\bottomrule
\end{tabular}%
}
\caption{Metrics of different methods on CIFAR-10, FashionMNIST and MNIST datasets.}
\label{tabel_metrics}
\end{table*}

\begin{figure*}[tb]
    \centering
    \begin{subfigure}{0.33\textwidth}
        \centering
        \includegraphics[width=\linewidth]{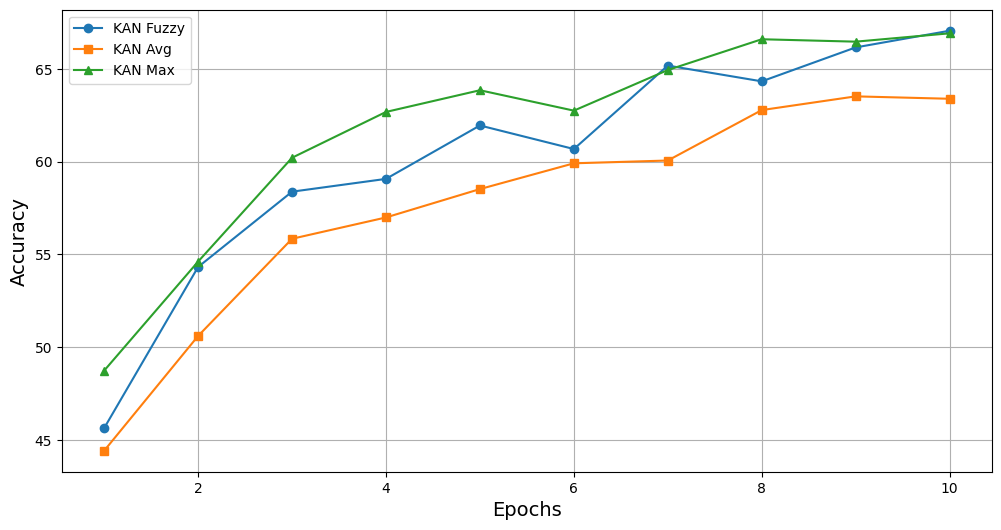}
        \caption{CIFAR-10}
    \end{subfigure}%
    \hfill
    \begin{subfigure}{0.33\textwidth}
        \centering
        \includegraphics[width=\linewidth]{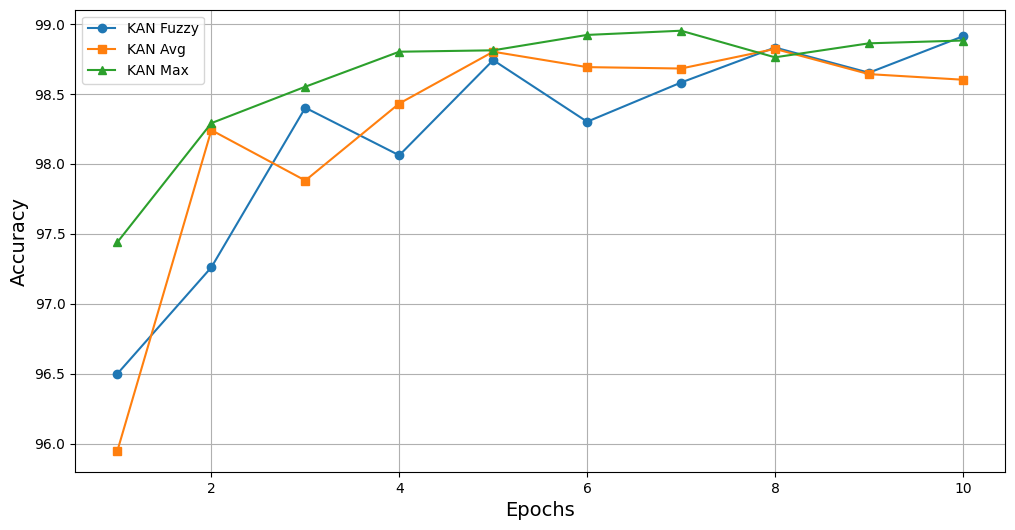}
        \caption{MNIST}
    \end{subfigure}%
    \hfill
    \begin{subfigure}{0.33\textwidth}
        \centering
        \includegraphics[width=\linewidth]{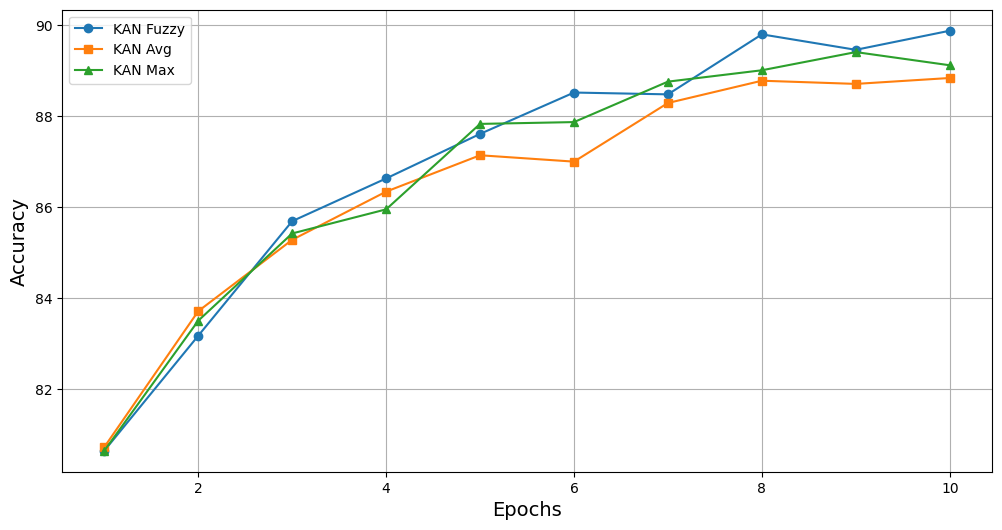}
        \caption{FashionMNIST}
    \end{subfigure}
    \caption{Accruacy of models over epochs with KAN on the last layer of CNN with different pooling layers.}
    \label{trend_plot}
\end{figure*}

\begin{figure*}[!h]
    \centering
    \begin{subfigure}{0.33\textwidth}
        \centering
        \includegraphics[width=0.8\textwidth]{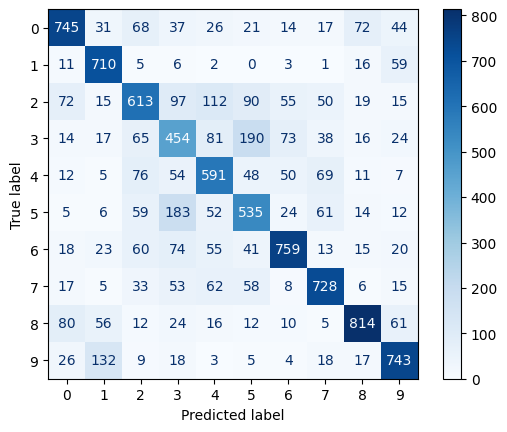}
        \caption{CIFAR-10 with Max Pooling}
    \end{subfigure}%
    \hfill
    \begin{subfigure}{0.33\textwidth}
        \centering
        \includegraphics[width=0.8\textwidth]{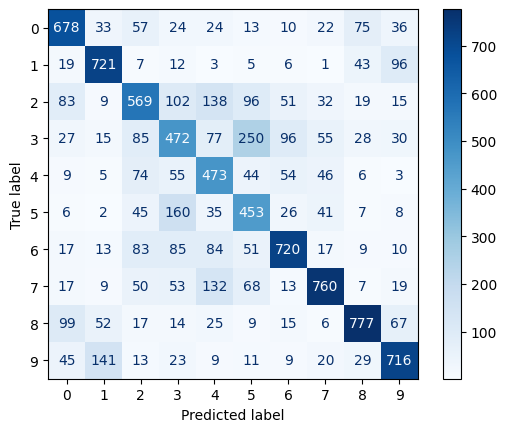}
        \caption{CIFAR-10 with Average Pooling}
    \end{subfigure}%
    \hfill
    \begin{subfigure}{0.33\textwidth}
        \centering
        \includegraphics[width=0.8\textwidth]{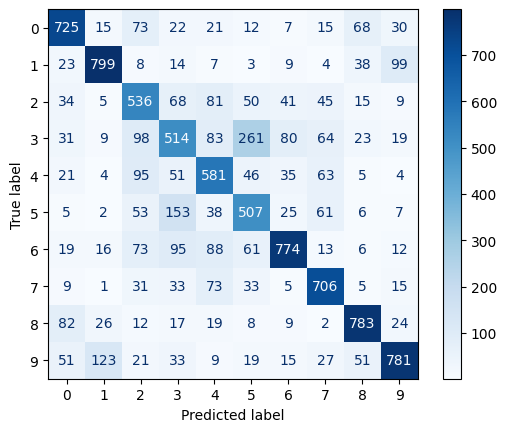}
        \caption{CIFAR-10 with Fuzzy Pooling}
    \end{subfigure}
    
    \hfill
    \begin{subfigure}{0.33\textwidth}
        \centering
        \includegraphics[width=0.8\textwidth]{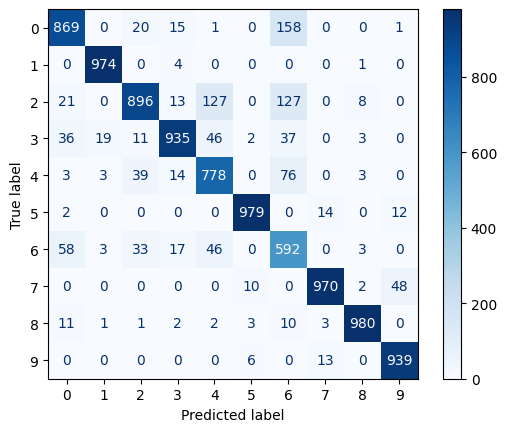}
        \caption{FashionMNIST with Max Pooling}
    \end{subfigure}%
    \hfill
    \begin{subfigure}{0.33\textwidth}
        \centering
        \includegraphics[width=0.8\textwidth]{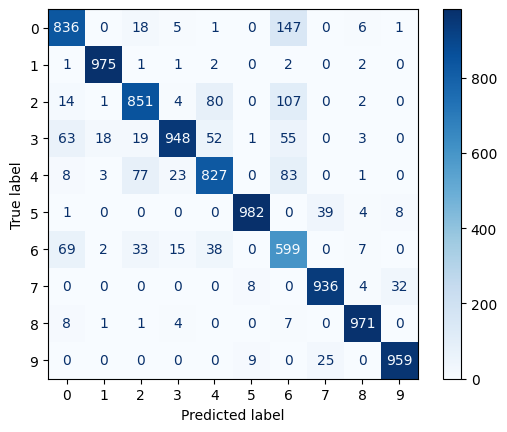}
        \caption{FashionMNIST with Average Pooling}
    \end{subfigure}%
    \hfill
    \begin{subfigure}{0.33\textwidth}
        \centering
        \includegraphics[width=0.8\textwidth]{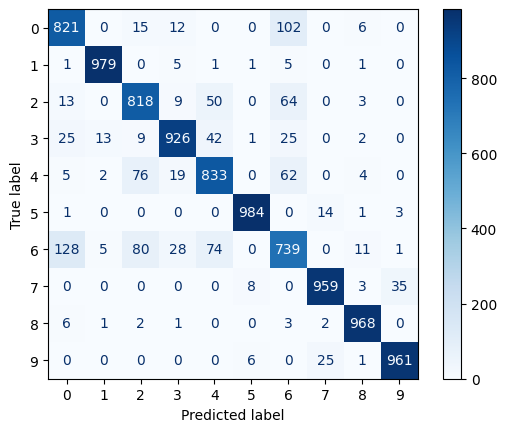}
        \caption{FashionMNIST with Fuzzy Pooling}
    \end{subfigure}
    
    \hfill
    \begin{subfigure}{0.33\textwidth}
        \centering
        \includegraphics[width=0.8\textwidth]{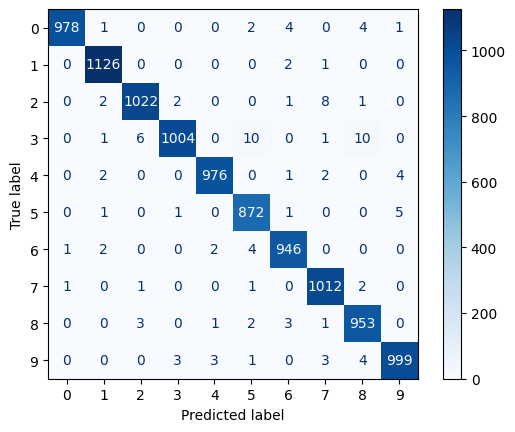}
        \caption{MNIST with Max Pooling}
    \end{subfigure}%
    \hfill
    \begin{subfigure}{0.33\textwidth}
        \centering
        \includegraphics[width=0.8\textwidth]{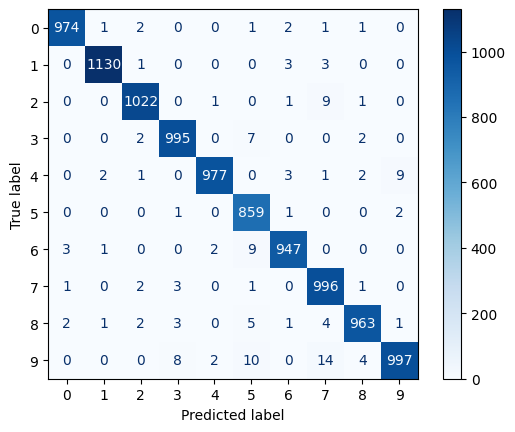}
        \caption{MNIST with Average Pooling}
    \end{subfigure}%
    \hfill
    \begin{subfigure}{0.33\textwidth}
        \centering
        \includegraphics[width=0.8\textwidth]{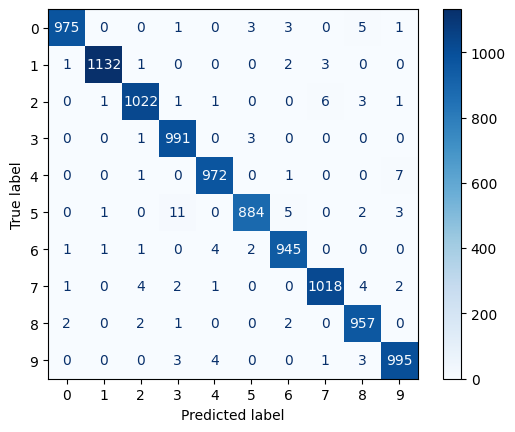}
        \caption{MNIST with Fuzzy Pooling}
    \end{subfigure}
    
    \caption{The multiclass confusion matrix of testing results for models with KAN on the last layer using different pooling methods over CIFAR-10, MNIST, and FashionMNIST datasets.}
    \label{full_confmat}
\end{figure*}


The developed models' results are illustrated in Table \ref{tabel_metrics}. As we can see, the proposed approach (CNN with Type-1 Fuzzy Pooling layers and KAN classification head) has the best performance, achieving the highest accuracy of 67.06\% on CIFAR-10, 89.88\% on FashionMNIST, and 98.91\% on MNIST. In comparison, the model with Average Pooling subsampling layers and KAN on the last layer yields lower accuracy values of 63.39\%, 88.84\%, and 98.60\% on CIFAR-10, FashionMNIST, and MNIST, respectively, highlighting the slightly better performance of the proposed approach. Models with Max Pooling as a subsampling layer have satisfactory results that almost match the performance of our approach. 

Figure \ref{trend_plot} illustrates the accuracy of models over epochs with KAN on the last layer with different pooling methods. The proposed model tends to have lower accuracy in the early epochs but steadily improves, eventually outperforming models with Average and Max Pooling in the later epochs.
Figure \ref{full_confmat} presents the multiclass confusion matrices for testing results of models with KAN on the last layer using different pooling methods on CIFAR-10, MNIST, and FashionMNIST datasets. Each subfigure (a-i) shows the performance of a specific pooling method and dataset combination, highlighting each model's accuracy and misclassification patterns.




\section{Conclusion}

The current paper introduces the novel architectural integration of KAN and fuzzy pooling in CNNs.
We conducted experiments on publicly available datasets, including CIFAR-10, MNIST, and Fashion-MNIST. As shown in Table \ref{tabel_metrics}, a modest accuracy improvement is consistent across all datasets.

Besides the performance improvement, our approach also brings progress toward explainable AI (XAI). The KAN offers more interpretable decision-making processes, and fuzzy pooling further enhances interpretability by effectively handling uncertainty and preserving essential features. In contrast, traditional CNNs with MLPs are often viewed as black-box models, providing little to no insight into their decision-making processes.


The current study has limitations. First, the dataset scope is limited, as the models were trained and tested on standard datasets that do not include complex images. Second, the architecture used, LeNet, is a relatively small CNN architecture.


As for future works,  we plan to perform experiments on larger datasets and use alternative CNN architectures. In addition, we aim to conduct experiments with different fuzzifications of the patches, such as using intuitionistic fuzzy sets or learnable boundaries for membership functions.
\section*{Acknowledgment}
This research has been funded by the Science Committee of the Ministry of Science and Higher Education of the Republic of Kazakhstan (Grant No. AP22786412)

\bibliography{references}

\end{document}